\documentclass[conference,10pt,letter]{IEEEtran}
\IEEEoverridecommandlockouts
\usepackage{cite}
\usepackage{amsmath,amssymb,amsfonts}
\usepackage{algorithmic}
\usepackage{graphicx}
\usepackage{textcomp}
\usepackage[caption=true,font=footnotesize]{subfig}

\usepackage{xcolor}
\usepackage{float}

\def\BibTeX{{\rm B\kern-.05em{\sc i\kern-.025em b}\kern-.08em
    T\kern-.1667em\lower.7ex\hbox{E}\kern-.125emX}}

\usepackage{multicol, multirow}
\usepackage{float}
    
\usepackage{xcolor}

\newcommand{\yd}[1]{{#1 }\color{black}}

\newcommand{\anish}[1]{{#1}\color{black}}

\begin{document}
\title{Enabling  On-Device Smartphone GPU based  Training:  Lessons Learned
}

\author{\IEEEauthorblockN{Anish Das}
\IEEEauthorblockA{\textit{University of Cambridge} \\
\textbf{\texttt{ad945@cam.ac.uk}}} 
\and
\IEEEauthorblockN{Young D. Kwon}
\IEEEauthorblockA{\textit{University of Cambridge} \\
\textbf{\texttt{ydk21@cam.ac.uk}}} 
\and
\IEEEauthorblockN{Jagmohan Chauhan}
\IEEEauthorblockA{\textit{University of Southampton} \\
\textbf{\texttt{j.chauhan@soton.ac.uk}}}
\and
\IEEEauthorblockN{Cecilia Mascolo}
\IEEEauthorblockA{\textit{University of Cambridge} \\
\textbf{\texttt{cm542@cam.ac.uk}}} 


}

\maketitle

\begin{abstract}

Deep Learning (DL) has shown impressive performance in many mobile applications. Most existing works have focused on reducing the computational and resource overheads of running Deep Neural Networks (DNN)  inference on resource-constrained mobile devices.  However, the other aspect of DNN operations, i.e. training (forward and backward passes) on smartphone GPUs, has received little attention thus far. To this end, we conduct an initial analysis to examine the feasibility of on-device training on smartphones using mobile GPUs. We first employ the open-source mobile DL framework (MNN) and its OpenCL backend for running compute kernels on GPUs. Next, we observed that training on CPUs is much faster than on GPUs and identified two possible bottlenecks related to this observation: (i) computation and (ii) memory bottlenecks. To solve the computation bottleneck, we optimize the OpenCL backend's kernels, showing 2x improvements (40-70 GFLOPs) over CPUs (15-30 GFLOPs) on the Snapdragon 8 series processors. However, we find that the full DNN training is still much slower on GPUs than on CPUs, indicating that memory bottleneck plays a significant role in the lower performance of GPU over CPU. The data movement takes almost 91\% of training time due to the low bandwidth. Lastly, based on the findings and failures during our investigation, we present limitations and practical guidelines for future directions.

\end{abstract}


\section{Introduction}

Deep Learning (DL) has gained much significance in the last decade due to its ability to model complex non-linear data efficiently. It has been successful in many domains, including computer vision, speech processing and natural language processing.  Existing works are increasingly moving towards supporting DL on resource-constrained mobile and Internet of Things (IoT) devices via optimizations designed to reduce the computation and memory requirements of Deep Neural Network (DNN) \textit{inference (i.e., forward pass)}~\cite{han_deep_2016,zhang_systematic_2018}. To name a few, weight quantization~\cite{han_deep_2016,jacob_quantization_2018}, pruning~\cite{zhang_systematic_2018}, vector quantization~\cite{martinez_permute_2021}, and new neural architectures~\cite{sandler_mobilenetv2_2018, ma_shufflenet_2018} have been proposed to make inference more efficient. However, the other side of the coin, DNN \textit{training (i.e., forward and backward pass)} \yd{on mobile systems such as smartphone's heterogeneous processors}, has been largely neglected~\cite{liu_performance_2019}. Moreover to address the increasing need of personalizing the deployed model according to the different user behaviors and changing environments effectively~\cite{kwon_exploring_sec21,servia-rodriguez_knowing_2021}, on-device training become essential in enabling pervasive computing~\cite{zhou_edge_2019}. Further, on-device training could provide personalization with lower latency and increased privacy because the data does not need to be shipped off to a server farm and the computation can be done on the device itself~\cite{chauhan_contauth_2020,kwon21_interspeech}. 

Modern devices, especially smartphones and smartwatches, are equipped with an array of accelerators such as GPU, which can realize the vision of enabling training on-device as it supports SIMD operations, i.e. the ability to apply the same operation on multiple data points in parallel, thereby expediting the process of learning by DNNs. However, popular DL frameworks such as TensorFlow~\cite{tensorflow} and PyTorch~\cite{pytorch} primarily focus on DNN training on a server or a cloud where abundant resources exist. Further, their lightweight counterparts for mobile devices (i.e., TensorFlow Lite and PyTorch Mobile) optimize the inference, leaving the training on mobile GPU not thoroughly optimized nor studied yet. Moreover, the recent development of mobile DL frameworks (e.g., MNN~\cite{jiang_mnn_2020}, Core ML) enables on-device training on smartphones. Such frameworks mainly focus on utilizing mobile CPU, which thereby could waste an opportunity to exploit heterogeneous hardware accelerators on-device.

In this paper, we investigate the feasibility of on-device training on smartphones using mobile GPUs for the first time: we evaluate to what extent operations can be optimized on a mobile GPU and provide lessons learned while doing the work.  For our purposes, we adopt to optimize the MNN framework (OpenCL backend) since MNN is open-sourced that can run on Android and iOS devices while Core ML is proprietary.
Initially, we executed MNN's OpenCL backend and ran the training with the MNIST dataset~\cite{mnist}. We observed that the CPU was much faster than the GPU. Thus, we proceeded to identify potential reasons for the poor performance and then improve the training performance on GPUs compared to CPUs.  Through our experiments on smartphones, we found out that computation-bound and memory-bound bottlenecks are the two primary reasons for the inefficiency of GPU on mobile platforms.
 
We firstly focused on solving computational bottlenecks. For this purpose, we introduce kernel optimizations for the matrix multiplication kernels in OpenCL to overcome the identified problems. 
After successfully implementing the optimizations, the optimized GPU kernels show speedups (40 and 70 GFLOPs) over CPU (15 and 30 GFLOPs) on One Plus 6 and One Plus 8 Pro. Nevertheless, we failed to gain improvements in the performance of full DNN training on the GPU. Upon further investigation, we discover that the memory bottlenecks hamper the overall goal of faster training for DNN using GPU (over CPU) on mobile devices. Our profiling experiments demonstrated that it takes about 10$\sim$15 $\times$ longer to copy data to and from the GPU than the actual computation time on all three devices. Therefore, even though we made significant improvements to the computation speed (optimizing matrix multiplications), ultimately, we see little improvement. 
We envision that solving this challenge will require the involvement of different disciplines, including hardware, architecture and systems that can enable faster training on mobile platforms using GPU.

The rest of the paper is organized as follows. Section~\ref{sec:preliminaries} provides the background knowledge of GPU architecture and the OpenCL programming model. Then, Section~\ref{sec:methodology} presents our kernel optimization techniques and study design to evaluate our GPU kernel implementation. After that, Section~\ref{sec:results} presents experimental results, and Section~\ref{sec:discussion} discusses the limitations, practical guidelines, and future work. Finally, we conclude the paper in Section~\ref{sec:conclusion}.

\section{Preliminaries}\label{sec:preliminaries}
This section provides a brief overview of the architecture of GPUs and the OpenCL programming model.


\subsection{GPU architecture}
\begin{figure}
    \centering
    \includegraphics[width=0.5\textwidth]{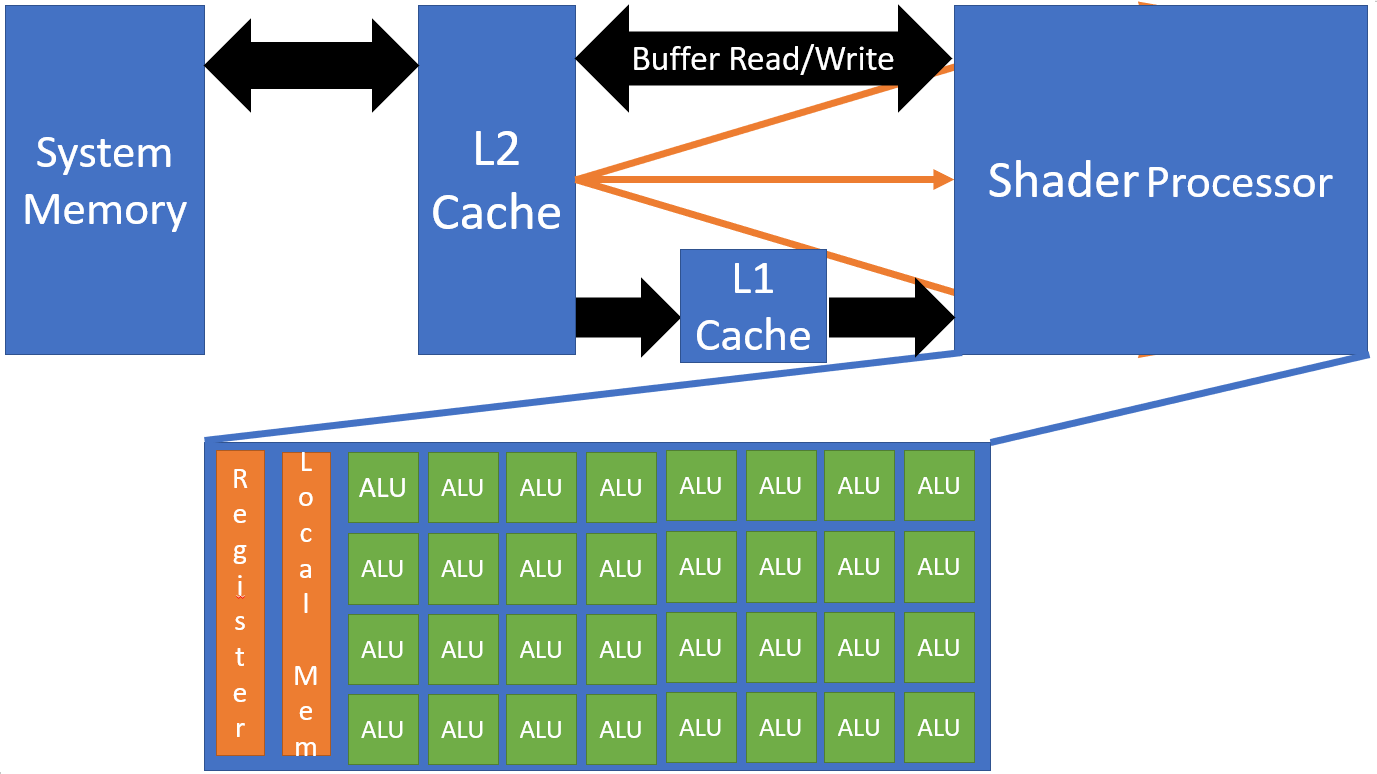}
    \caption{High Level Overview of the Adreno GPU architecture for OpenCL}
    \label{fig:gpu-fig}
\end{figure}

\anish{To explain the architecture of the smartphone GPUs, we use a high level diagram of Adreno GPUs from the perspective of programming in OpenCL. As shown in Figure \ref{fig:gpu-fig}, the GPU is comprised of a set of Shader Processors (SP). The SPs are comprised of many ALUs which execute in lock-step. Each ALU has different memory spaces available to it with drastically different latencies. The memory spaces on the SP include registers, which is very fast, followed by Local Memory, accessible to all ALUs. Outside the SP, there is the L1 cache which is read-only and is used for Texture Processing, the L2 cache where the Buffers/Image datatypes are stored and finally, the global memory (system memory).         } 

\subsection{OpenCL programming model}
OpenCL is a framework for writing computationally-oriented kernels to run on heterogeneous computing devices. It has a controller-device execution model, a host process usually on CPUs, creating and managing tasks to be run on GPUs of a computing device. The computing device can be any form of a hardware accelerator such as GPU, Digital Signal Processor (DSP), or Field Programmable Gate Arrays (FPGA).
The host's responsibility is to compile the kernel code for the target device and then send an execution task to the command queue. The command queue is responsible for scheduling tasks onto GPUs, and it runs on the host process (sometimes asynchronously). 


The four main parts of the OpenCL programming model are: 
\begin{enumerate}
    \item Work Item: This can be regarded as an independent thread that runs the kernel program. It is called fiber in the Qualcomm Adreno GPUs. 
    \item Work Group: A collection of threads/work-item that execute in lockstep are called a workgroup. These will be scheduled onto an SP (from Figure \ref{fig:gpu-fig}). The workgroup has a limit due to the fixed number of SP. 
    \item Local Size: The local sizes define the dimension of each workgroup. In OpenCL, up to a three-dimensional arrangement of work-item is possible. The workgroup size is the product of the dimensions. 
    \item Global Size: Global size defines the total number of threads that need to be run. It can also be up to 3 dimensional, and there is no limit.
\end{enumerate}


The memory hierarchy in OpenCL closely tracks the memory arrangement on GPUs as described previously. It has four types of memory as follows.
\begin{enumerate}
    \item Private Memory: The registers generally account for this memory. It is only accessible to each thread, i.e. each SP has its private memory and any intermediate variables created are stored here. 
    \item Constant Memory: This is a fast read-only memory that can also be written to by the host, generally used for textures. In the Qualcomm guidebook\footnote{{https://developer.qualcomm.com/qfile/33472/80-nb295-11\_a.pdf}}, the constant memory is stored in the L1 cache.
    \item Local Memory: This is a very fast memory that is cached on the SP. It is the memory that is shared among all the threads in a workgroup. 
    \item Global Memory: The largest and the slowest memory shared by all of the threads.
\end{enumerate}

\begin{table*}[h]
    \begin{center}
    \begin{tabular}{|l|l|l|l|}
\hline

Name & System-on-Chip (Qualcomm) & CPU & GPU \\ \hline
 &  & Octa-core (1x2.4 GHz Kryo 475 Prime \& &  \\
One Plus Nord & Snapdragon 765G 5G (7 nm) &  1x2.2 GHz Kryo 475 Gold \& & Adreno 620 \\
 &  &  6x1.8 GHz Kryo 475 Silver) &  \\ \hline
 &  & Octa-core   &  \\
One Plus 6 (OP6) & Snapdragon 845 & 4×2.8 GHz Kryo \& & Adreno 630 \\
 &  & 4×1.8 GHz Kryo &  \\
\hline
 &  & Octa-core (1x2.84 GHz Kryo 585 \& &  \\
One Plus 8 Pro (OP8) & Snapdragon 865 5G (7 nm+) & 3x2.42 GHz Kryo 585 \& & Adreno 650 \\
 &  & 4x1.8 GHz Kryo 585) &  \\ \hline

 \multicolumn{4}{c}{\phantom{end-of-table}}
\end{tabular}
\end{center}
    \caption{The employed devices and their specifications.}
    \label{tab:devices-used}
\end{table*}

\section{Methodology}\label{sec:methodology}


In this section, we first present how we optimize the GPU kernels of MNN to maximize the parallel computations of matrix multiplication (\S\ref{subsec:method kernel optimizations}). We then describe the experimental setup in \S\ref{subsec:method experimental setup}. After that, we explain the study design for evaluating our optimized GPU kernels compared to the na\"ive GPU kernel and CPUs in \S\ref{subsec:method study design}.

\subsection{Kernel Optimizations}\label{subsec:method kernel optimizations}

Matrix multiplication operation is one of the most crucial tasks in the DL computation cycle, and hence we optimize this operation using kernel optimizations in this subsection. We discuss the improvements made to a na\"ive kernel that lead to the improved performance. For this step, we tried optimization strategies for Single-precision matrix multiplication (SGEMM). We mainly focus on and describe two types of optimizations: Tiling and Vectorization, as they provide the best performance amongst all the available optimization strategies in our experiments. 


\begin{figure}
    \centering
    \includegraphics[width=0.5\textwidth]{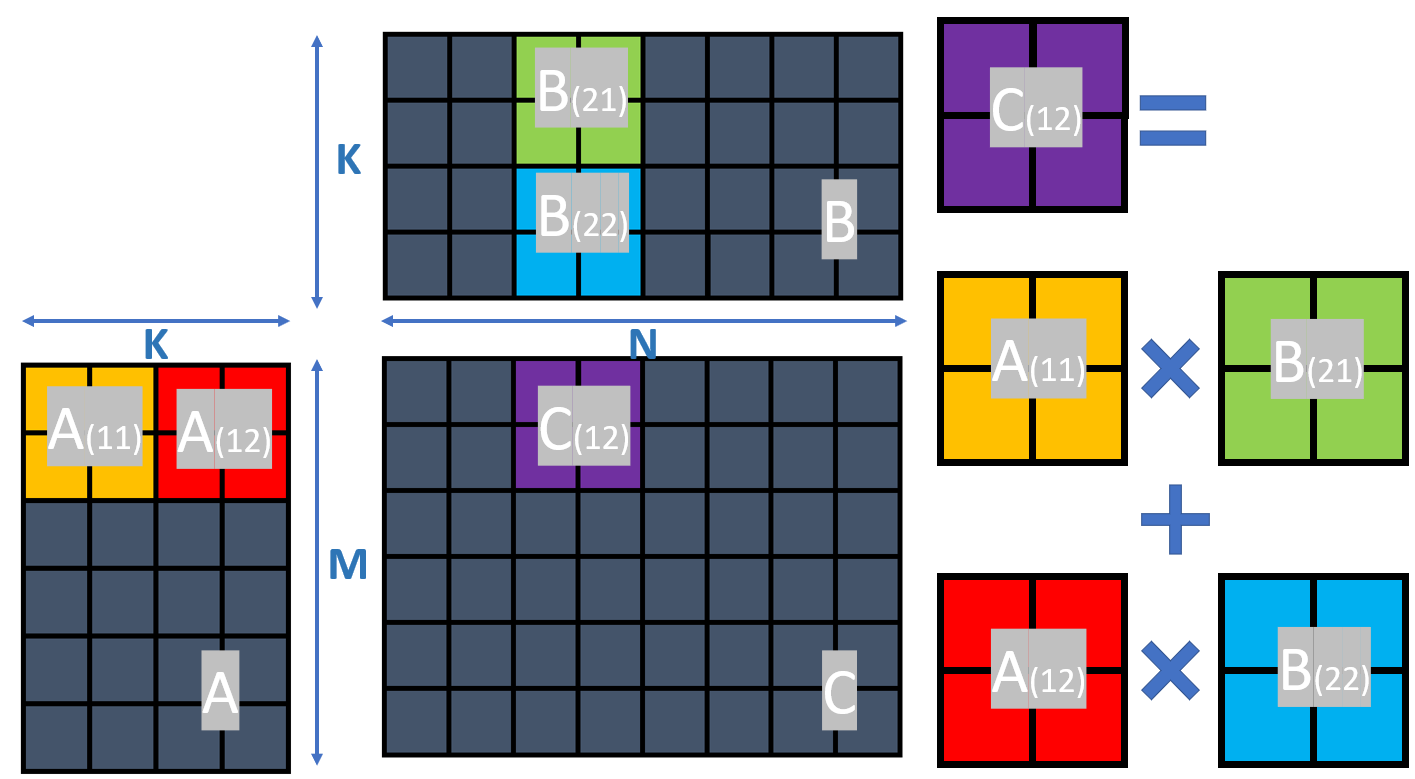}
    \caption{
    The visualization of the tiling kernel optimization. 
    A work group is assigned to calculate the block $C_{(12)}$. During the calculation, the kernel first loads blocks $A_{(11)}$ and $B_{(21)}$ and proceeds to calculate the partial solution. Then it loads the next blocks $A_{(12)}$ and $B_{(22)}$ and uses them to compute the other half of the partial solution. 
    }
    \label{fig:tiling}
\end{figure}

\subsubsection{Na\"ive Kernel}
The na\"ive kernel assigns one work item to each output position in the output matrix C in Figure 2. It then iterates over the common dimension in [MxK] times [KxN]. It then iterates over all the K values in the rows of A and columns of B. Thus, it is a very inefficient kernel because threads next to each other on a workgroup will request the same row/column. Moreover, it does not take advantage of the spatial and temporal locality related to the computation of the kernel. 
Each workgroup executes in lockstep, and if all work items continuously request indices from the input matrices, it would lead to clashes, consequently, the data would have to be fetched from the global memory. 
Two simple optimizations showing the best performance improvement are described next.

\subsubsection{Vectorization}
OpenCL provides functionality that allows for vector data types to be used in the kernel code. Vector data types are, as the name suggests, data-types arranged in a contiguous block of size 2, 4, 8 or 16. It is known that memory is stored in cache lines (32 bytes long) for mobile GPUs. When data is read in, the entire line is read in, and therefore, when we load the value at some index, the data at the neighboring positions are also loaded in (`Spatial Locality'). When we use vector data types, we can exploit the spatial locality as the load and store operations very closely mirror the actual operations in the hardware. Through experimentation, we found that a vector size of four is optimal. 

In addition, the vectorization has a secondary benefit of having a lower workgroup count because each work item is now responsible for multiple indices in the output matrix. With a lower workgroup count, a larger number of kernels can be scheduled to execute in each warp and consequently boost the overall throughput. 


\subsubsection{Tiling}

For tiling, we leverage the local memory (shared among the work items in a workgroup) to cache a portion of the input matrices and then reuse them for each work item. This step is done iteratively, and it reduces the number of load operations by a factor equal to the local size in each dimension. 
Thus, the values at each index need to be loaded only a few times.  In this step, we create a  matrix in local memory which has the same shape as the workgroup. Next, each workgroup is assigned a task to load a specific index, and each index is loaded concurrently. Then, the intermediate results are calculated for the cached matrix portion for each index, respectively. Figure~\ref{fig:tiling} visualizes the tiling kernel optimization.

\begin{figure*}
    \centering
    \includegraphics[width=\textwidth]{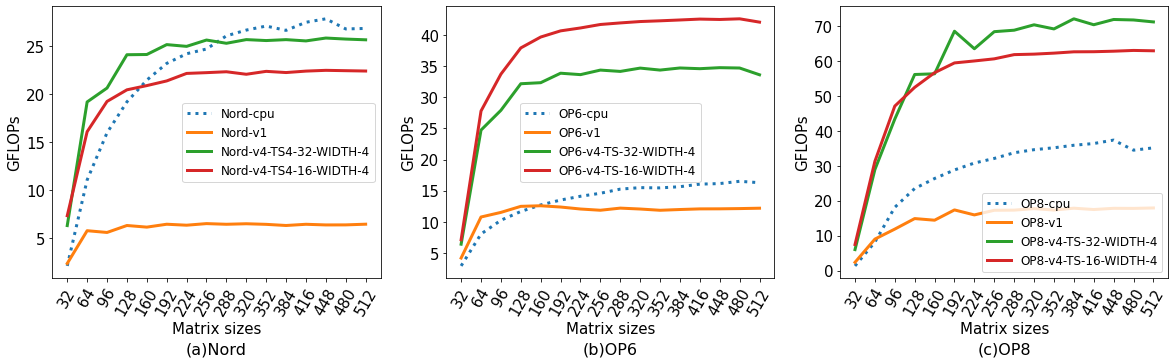}
    \caption{Profiling results of the optimized OpenCL backend kernels for the matrix multiplication operation.
    }
    \label{fig:matmul-profile-output}
\end{figure*}

\subsection{Experimental Setup}\label{subsec:method experimental setup}
\subsubsection{Hardware}
In our experiments, we use three devices: One plus Nord, One Plus 6, and One Plus 8 Pro. The details of the mobile processors (i.e., CPU and GPU present on smartphones) are shown in Table~\ref{tab:devices-used}. The One Plus 6 is the oldest phone and was released in May 2018. The other two devices were released in 2020, with the One Plus 8 released as a flagship and the Nord released as a cheaper alternative. Therefore, the Nord has the older version of Adreno GPU but has a newer CPU compared to the One Plus 6. 

\subsubsection{Profiling Tools} We use two profiling tools: (1) GPU timer and (2) OpenCL profiler. Firstly, the GPU timer is a tool present in the OpenCL API that allows an Event object to be attached to a kernel invocation. The Event object is a bookkeeping tool that tracks the time it takes to process the kernel call, i.e. time it takes to add the kernel call onto the command queue. It also tracks the time the command spends in the queue and its time to execute on the GPU. The GPU timer is used in the profiling experiments to accurately measure the latency to run the kernels on the GPU. 

Secondly, the MNN framework provides an option to use the OpenCL profiler, which tracks all the operations being performed and reports their time to execute. It uses the GPU timer for execution time and further reports its time to copy data to and from the GPU.

\subsection{Study Design}\label{subsec:method study design}

\subsubsection{Profiling Kernel Optimizations}
To profile the matrix multiplication kernels, we ran the matrix multiplication operation on square matrices of various sizes starting from [32$\times$32] to [512$\times$512] in steps of 32. The operation was run multiple times by queuing the operation multiple times with the same arguments. Keeping the arguments the same ensures that we do not incur the cost of moving the data to/from the GPU. We ran a few warm-up operations since the execution speed usually takes some time to reach the maximum speed. We ran the matrix multiplication for 500 hot runs and track the execution time following the warm-up steps. After collecting the execution times, the GFLOPS value is calculated using the standard formula: FLOPS = $2n^3$ / (avg. execution time)

\subsubsection{Profiling Back Propagation}
After we confirmed that the matrix multiplication operation on the GPU is faster than the CPU on all the tested smartphones (after kernel optimizations), the next step was to incorporate these new kernels into the existing MNN training API. When we ran the experiments on the training throughput, we employed the MNIST dataset \cite{mnist} \yd{since it is a commonly used dataset for an initial benchmark.}
In the context of training, the operation that is most important and the one that takes the most amount of time is the backward propagation. From our experiments, we report the backward propagation time on all the devices for each of the models at a fixed batch size of 32. Furthermore, we also report the ratio of execution time on the GPU over the execution time on CPU. The ratios are meant to reduce the cognitive load when trying to compare the CPU vs GPU. 
A simple feed-forward network model with four layers (and nodes) as: $784 \rightarrow 256 \rightarrow 128 \rightarrow 16$ was used to run the experiments. The time  taken for each backward pass was measured after discarding the first few warm-up runs. Our optimized kernels did not help improve GPU's performance over CPU for training and hence we investigated further with different model architectures to find the underlying cause. These details are described next.

\textbf{Model Architecture.} Having failed at beating the CPU in the experiments devised above, we move on to trying out different architectures to test how size affects the latency. Instead of finding a new dataset to meet our input size requirements, we chose to use randomized inputs (matrices filled with random numbers) because we are only interested in the average time it takes for the forward and backward pass. 

We used feed-forward neural networks with the following shapes in our experiments: 
\begin{itemize}
    \item Model1: $16 \rightarrow 16$
    \item Model2: $16 \rightarrow 16 \rightarrow 16 \rightarrow 16$
    \item Model3: $256 \rightarrow 256$
    \item Model4: $784 \rightarrow 256 \rightarrow 128 \rightarrow 16$
\end{itemize}

\begin{figure*}[h]
    \centering
    \includegraphics[width=0.95\textwidth]{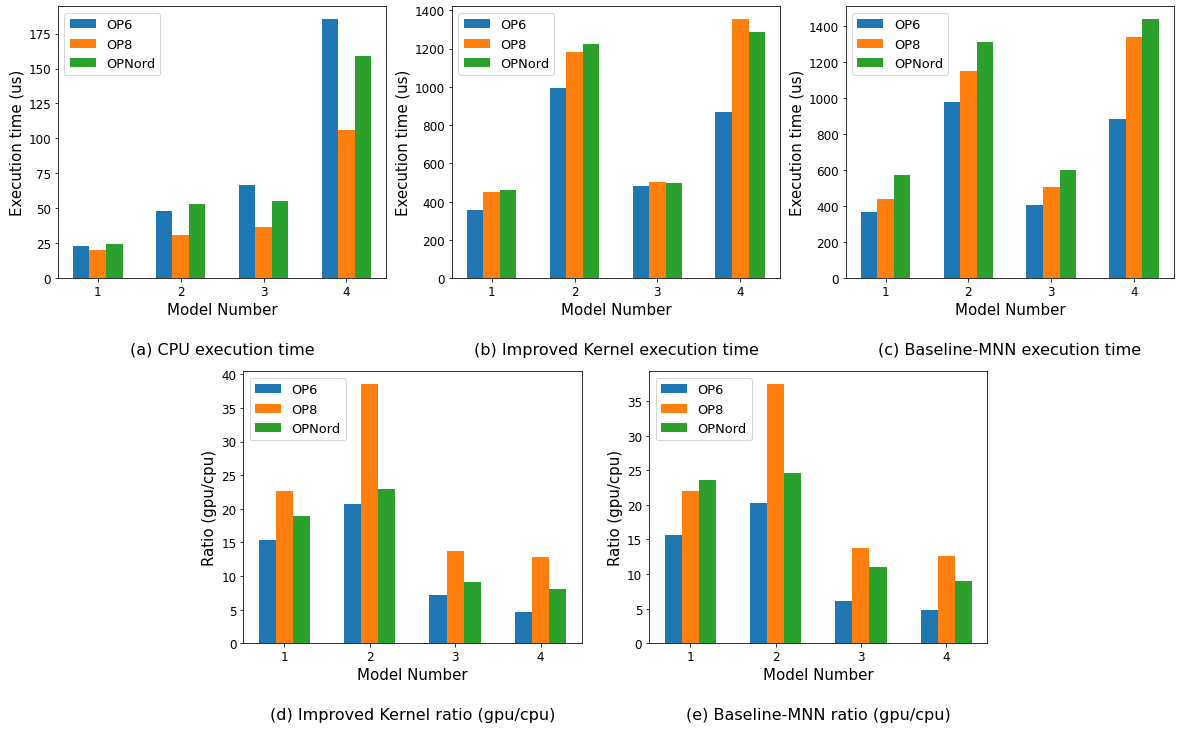}
    \caption{The back propagation time on (a) CPU, (b) our improved kernel GPU implementation, and (c) the baseline MNN GPU implementation for all the three devices. The ratio of the execution time on GPUs vs. CPUs of (d) our improved kernel implementation and (e) the baseline MNN GPU implementation. 
    }
    \label{fig:exec-and-ratio}
\end{figure*}

\subsubsection{Profiling Compute and Memory Operations for Training} 
After we observed little change in the ratios (backward propagation time on GPU/CPU) for the baseline-MNN kernels and the improved kernels, we devised a final experiment to calculate the time taken for moving the matrices onto the GPU memory and the time spent on actual computation. We employed the OpenCL Profiler, part of the MNN framework, to measured the times for all models mentioned before.

\section{Results}\label{sec:results}

\subsection{Profiling Kernel Optimizatons}\label{subsec:kernel optimization}


We now present the profiling results of kernel optimization for the matrix multiplication operation. Figure~\ref{fig:matmul-profile-output} shows the speed of different matrix multiplication kernels on CPU and GPU in terms of GFLOPS. To begin with, the naive GPU kernel is slower than CPUs on all the three smartphones. Then, our optimized GPU kernels improve the GFLOPS remarkably from 5, 10, 15 GFLOPS to 25, 40, 70 GFLOPS on One Plus Nord, One Plus 6, and One Plus 8 Pro, respectively. In case of One Plus Nord which is based on a Snapdragon 765 processor, our optimized GPU kernel implementation shows similar performance as CPU. But, for One Plus 6 and One Plus 8 Pro which are based on Snapdragon 8 series processors equipped with better mobile GPUs, our optimized GPU kernels start to outperform  CPU. The  execution speed on GPU is at least 2x faster than CPU on both the devices. 


\subsection{Profiling  Back Propagation}


After we obtained better performance on GPU than CPU for matrix multiplication, we moved to profile the back propagation using our optimized kernels. Figure 4 shows the overall latency of the backward pass (a-c) and the ratio of running time with GPU over CPU (d-e).
The comparison of the execution time between the smaller and larger models (Model1 vs. Model3 and Model2 vs. Model4) is roughly similar. This shows that the computation on the GPU is very fast.
However, the deeper models (Model2 and Model4) show the longer overall back propagation latency on GPUs compared to the shallow models (Model1 and Model3). This result indicates that the data movement is the major bottleneck that hinders the overall performance of GPU. 
The data movement bottleneck affects the deeper models more because each layer in the feed-forward neural networks requires new matrix multiplication operations, and each operation requires additional data movement.
For example, the back propagation time of the deeper models with three layers is about 3$\times$ longer than the shallow models. Thus, the back propagation time grows linearly with the number of layers. 
In addition, the higher ratio means CPU is faster than GPU. First of all, we find that the ratios for the baseline MNN kernel and optimized GPU kernel have little to no improvement in back propagation latency as optimized GPU kernel does not show consistent improvement over the baseline as in Figures 4d and 4e. We attribute this result to the fluctuations in the time taken for data movement since the improvement in computation times are very consistent as demonstrated in \S\ref{subsec:kernel optimization}. 
However, there is a rise in relative performance for the larger models: Model3 and Model4, which have small ratio values as larger models benefit more from computation parallelism than smaller models (i.e., Model1 and Model2). This result reaffirms that the major factor of low performance of back propagation latency on mobile GPUs is the memory bound.

\subsection{Profiling Compute and Memory Operations for Training} 

To confirm that the problem lies in the memory bandwidth between the CPU and GPU, we estimate the time spent on computation versus time spent on data movement between the main memory and the GPU memory. We use the OpenCL profiler in MNN and check the average the time spent on copying and the time spent on kernel execution during the training.  The results indicate that the computation time accounts for only about 9\% of the total training time in all three of the phones employed. 
Therefore, these results further confirm that the primary factor of slower training time on mobile GPUs is the memory bottleneck rather than the computation bottleneck.



\section{Discussion and Future Directions}\label{sec:discussion}

\yd{Despite the best of efforts, this work has a few shortcomings and also provides practical suggestions for future work.
First, we need the minimum matrix dimension to be 16 for all the matrices involved (inputs and outputs). Although it does not affect the overall profiling, it would be interesting to accommodate arbitrary matrix sizes.
Further, since the focus was on accelerating matrix multiplication operation and thus enabling feed-forward network models, this work has not investigated convolutional neural networks. However, since convolutional layers are computationally more expensive than fully-connected layers (i.e., our target to improve in this work) as analyzed in~\cite{alzantot_rstensorflow_2017} and the real bottleneck of on-device training is memory bound as in Section 4, analyzing and improving computationally expensive convolutional layers can be a potential future direction. 
Our finding, memory bottleneck issue, also suggests investigating how to enable on-device training with memory optimization in terms of model, optimizer, and the activation~\cite{sohoni_low-memory_2019}.
}


\section{Conclusion}\label{sec:conclusion}

We have investigated the feasibility of on-device training on resource-constrained mobile GPUs with three different smartphones based on the OpenCL backend of the mobile DL framework, MNN. First, we started with the observation that training on mobile GPUs is much slower than that on mobile CPUs. We then identified two obstacles regarding computation and memory. Through extensive GPU kernel optimization to maximize the computation parallelism, we improved the throughput of GPUs over CPUs, resolving computation bound problem. However, our experimental results show that overall training time on GPU remains higher than CPU, indicating that memory bound is a major factor contributing to the inefficient training on mobile GPUs. Though we failed to accelerate the full training on GPUs over CPUs, our research findings could provide practical guidelines for further development of on-device training and better utilization of heterogeneous mobile processors with CPUs and GPUs in the near future.

\section*{Acknowledgment}
This work is supported by a Google Faculty Award 2019 and by Nokia Bell Labs through their donation for the Centre of Mobile, Wearable Systems and Augmented Intelligence to the University of Cambridge


\bibliographystyle{IEEEtran}
\bibliography{refs}

\end{document}